\documentclass[10pt, a4paper]{article}
\usepackage{lrec2022} % this is the new LREC2022 Style
\usepackage{multibib}
\newcites{languageresource}{Language Resources}
\usepackage{graphicx}
\usepackage{tabularx}
\usepackage{soul}
\usepackage{titlesec}
\titleformat{\section}{\normalfont\large\bfseries\center}{\thesection.}{1em}{}
\titleformat{\subsection}{\normalfont\SmallTitleFont\bfseries\raggedright}{\thesubsection.}{1em}{}
\titleformat{\subsubsection}{\normalfont\normalsize\bfseries\raggedright}{\thesubsubsection.}{1em}{}
\renewcommand\thesection{\arabic{section}}
\renewcommand\thesubsection{\thesection.\arabic{subsection}}
\renewcommand\thesubsubsection{\thesubsection.\arabic{subsubsection}}
\usepackage{epstopdf}
\usepackage[utf8]{inputenc}

\usepackage[hidelinks]{hyperref} % Edited by harish, remove if reviewers complain.
\usepackage{xstring}

\usepackage{color}

\usepackage{multirow}
\usepackage[inline]{enumitem}
\usepackage{scrextend}
\usepackage{array}
\usepackage{arydshln}
\usepackage{graphicx}
\usepackage{hhline}
\usepackage{todonotes}
\usepackage{amsmath}

\usepackage{arydshln}
\title{Sample Efficient Approaches for Idiomaticity Detection}

\name{Dylan Phelps, Xuan-Rui Fan, Edward Gow-Smith, \\[0.15cm]
{\large \textbf{Harish Tayyar Madabushi, Carolina Scarton, Aline Villavicencio}}
\\[-0.1cm]
}

\address{
Department of Computer Science, \\
University of Sheffield \\
United Kingdom \\
\texttt{\small \{drsphelps1, lhsu1, egow-smith1, h.tayyarmadabushi, c.scarton, a.villavicencio\}} \\[-0.05cm]
\texttt{\small @sheffield.ac.uk}
\\[-0.2cm]
\\
}

\vspace{0.2cm}

\abstract{
Deep neural models, in particular Transformer-based pre-trained language models, require a significant amount of data to train. This need for data tends to lead to problems when dealing with idiomatic multiword expressions (MWEs), which are inherently less frequent in natural text. As such, this work explores \emph{sample efficient} methods of idiomaticity detection. In particular we study the impact of Pattern Exploit Training (PET), a few-shot method of classification, and BERTRAM, an efficient method of creating contextual embeddings, on the task of idiomaticity detection. In addition, to further explore generalisability, we focus on the identification of MWEs not present in the training data. Our experiments show that while these methods improve performance on English, they are much less effective on Portuguese and Galician, leading to an overall performance about on par with vanilla mBERT. Regardless, we believe sample efficient methods for both identifying and representing potentially idiomatic MWEs are very encouraging and hold significant potential for future exploration.
\\ \newline  }

\begin{document}

\maketitleabstract

\section{Introduction and Motivation}
The handling of idiomaticity is an important part of natural language processing, due to the ubiquity of idiomatic multiword expressions (MWEs) in natural language \cite{sag2002multiword}. As such, it is an area where the performance of state-of-the-art Transformer-based models has been investigated \cite{yu2020assessing,garcia2021probing,nandakumar-etal-2019-well}, with the general finding being that, through pre-training alone, these models have limited abilities at handling idiomaticity. However, these models are extremely effective at transfer learning through fine-tuning, and thus are able to perform much better on supervised idiomatic tasks \cite{fakharian2021contextualized,kurfali-ostling-2020-disambiguation}, where significant amounts of labelled data is provided. 

Unfortunately, individual MWEs tend to occur infrequently in natural text, making it harder to train models to capture the idiomatic meaning due to the lack of available training data. As such it is important to be able to find methods of identifying potentially idiomatic MWEs using relatively less data. To address this question we focus on \emph{sample efficient} methods for the task, taking two perspectives. The first is an evaluation of a few-shot method on the task of zero-shot idiomaticity detection. In particular we evaluate Pattern Exploit Training (PET)~\cite{schick-schutze-2021-exploiting}, which has been shown to be an effective few-shot method on other tasks~\cite{schick-schutze-2021-just}. The second is an evaluation of the effectiveness of better representations of MWEs, created using a sample efficient strategy, namely BERTRAM~\cite{schick-schutze-2020-bertram}. Both of these are explored in the zero-shot context, where training data does not include MWEs present in the test data. So as to ensure reproducibility and to enable others to build upon this work, we make the programme code and models publicly  available\footnote{\href{https://github.com/drsphelps/idiom-bertram-pet}{https://github.com/drsphelps/idiom-bertram-pet}}.

\subsection{Research Questions and Contributions}

Given the need for sample efficient methods when dealing with idiomaticity, this work is aimed at exploring the following questions: 

\begin{itemize}
    \item How effective are few-show methods on the task of zero-shot idiomaticity detection? In particular we evaluate Pattern Exploit Training (PET)~\cite{schick-schutze-2021-exploiting}, which has been shown to be an effective few-shot method on other tasks~\cite{schick-schutze-2021-just}. 
    \item Given that prior work has shown pre-trained language models do not adequately capture multiword expressions, in particular those which are idiomatic, how effective is improving their representations on the task of detecting idiomaticity? In particular, we use BERTRAM~\cite{schick-schutze-2020-bertram} as a sample efficient strategy for creating representations of MWEs.
\end{itemize}

From our experiments, we find that both BERTRAM and PET are able to outperform mBERT \cite{devlin-etal-2019-bert} significantly on the English portion of the test data, which is a promising result. However, both of these models perform worse overall due to their significantly lower performance on Portuguese. We explore potential reasons for this poor performance on non-English languages: for PET our patterns are all in English and a multilingual model is used instead of a language specific one. However, an error analysis (Section \ref{sec:error-analysis}) suggests that these are not the reasons for the lower performance on non-English languages. In BERTRAM, however, a monolingual model is used for each language which might have contributed to the drop in performance. We believe that these results point to the need for further exploration in languages other than English.%, especially those which are low resource. 

Additionally, our exploration using BERTRAM is, to the best of our knowledge, the first work to explore the relation between the representation and detection of idiomaticity.

The rest of this paper is structured as follows. We begin in Section \ref{section:related-work} by presenting a quick overview of work related to MWE identification, before presenting more details of the methods we make use of in this work. We then provide an overview of the data and task we use for our evaluation in Section \ref{section:semeval}, before presenting the methods in Section \ref{section:methods}. We then present our results and a discussion of what these results imply in Section \ref{section:results}, before concluding in Section \ref{section:conclusion}.

\section{Related Work} \label{section:related-work}

Despite idiomaticity detection being a problem that has been widely explored \cite{constant-etal-2017-survey}, the impact of better MWE representations, especially within contextualised models, has not been well studied. To this end we use BERT for Attentive Mimicking (BERTRAM) \cite{schick-schutze-2020-bertram}, which has been shown to perform well on idiom representation tasks \cite{phelps-semeval}, to evaluate the effect idiom representations have on detection. Additionally, we apply a few-shot learning technique Pattern Exploit Training (PET) \cite{schick-schutze-2021-exploiting}, to assess whether the relatively new paradigm of few-shot learning can be applied to this task successfully.

\begin{table*}[ht!]
\begin{center}
\begin{tabular}{|c|c|c|c|}
      \hline
      Pattern Number & Pattern & Literal Token & Idiom Token \\
      \hline
      P1 & X: \_\_\_\_ & literal & phrase \\
      \hline
      P2 & (\_\_\_\_) X & literal & phrase \\
      \hline
      P3 & X. [IDIOM] is \_\_\_\_ literal.& actually & not \\
      \hline
      P4 & X. \_\_\_\_, [IDIOM] is literal. & yes & no \\
      \hline
      P5 & X. [IDIOM] is \_\_\_\_ [IDIOM]\textsubscript{2} & actually & not \\
      \hline
\end{tabular}
\label{table:PVPs}
\caption{Pattern Verbaliser Pairs used in the task. X represents the example sentence, [IDIOM] is the idiom found in the example, and [IDIOM]\textsubscript{2} represents the n\textsuperscript{th} component word of the idiom}
\end{center}
\end{table*}

\subsection{PET} % Dylan
PET \cite{schick-schutze-2021-exploiting,schick-schutze-2021-just} is a semi-supervised training method that improves performance in few-shot settings by integrating task descriptions into examples. 

A Pattern is used to map each example into a cloze-style question with masked out tokens, for example \textit{`X. It was [MASK]'}, where X is the input example, could be used for a sentiment classification task. A Verbaliser maps the task classes into outputs from the masked language model (MLM), for example positive/negative labels map to the words `good'/`bad' in the MLM's vocabulary (label tokens), and is combined with the pattern to form a Pattern Verbaliser Pair (PVP). The probability of each class is then calculated using softmax over the logits for each label token.

For each PVP, an MLM can be fine-tuned on the small amount of labelled data. Knowledge is distilled from multiple PVPs by combining the predictions on the unlabelled data and using it as a larger labelled dataset to train another classifier. This allows for multiple patterns and verbalisers to be used without having to choose the best performer for each task, which may also change depending on the data split.

\subsubsection{iPET}
iPET \cite{schick-schutze-2021-exploiting} is a variation where each PVP's model is trained iteratively using a gradually increasing training set made up of labelled examples from another model's predictions in the previous iteration. Despite using the same PVPs and MLMs, iPET has been shown to improve the performance on a number of tasks \cite{schick-schutze-2021-just}.

% \subsubsection{ADAPET}
% A Densely-supervised Approach to Pattern Exploiting Training (ADAPET; \cite{tam-etal-2021-improving}) modifies the training objbective for PET via two methods, such that no task-specific unlabelled data is required. The first method, Decoupling Label Loss, changes the loss function to incorporate information from all tokens in the vocabulary not just from the class tokens defined by the verbaliser. This is done by calculating the probability of each class using softmax over all the tokens, rather than just the class tokens. Secondly another task is added which asks the model to predict other words in the context given one of the labels. If the correct label is passed the model must predict the original masked token, if the incorrect label is passed the model will be penalised for predicting the original.

% ADAPET was shown to have higher performance than PET and iPET on the SuperGLUE benchmark, despite not using any unlabelled data and only 32 examples for each task.

\subsection{BERTRAM} % Dylan

BERTRAM \cite{schick-schutze-2020-bertram} is a model for creating embeddings for new tokens within an existing embedding space, from a small number of contexts. To create an embedding for a token with a number contexts, a form embedding is first created using embeddings trained for each of the n-grams in the token. This form embedding is then passed as an input, alongside the embeddings for words in the context, into a BERT model. An attention layer is then applied over the contextualised embedding output from BERT for each context to create the final embedding for the token.

The model is trained using embeddings for common words as `gold standard' embeddings, with the distance from the embedding created by the model and the `gold standard' embedding being used as the loss function.

\section{Dataset and Task Description} \label{section:semeval}
In evaluating the models presented in this work we use the Task 2 of SemEval 2022: Multilingual Idiomaticity Detection and Sentence Embedding \cite{tayyarmadabushi-etal-2022-semeval}. This task aims at stimulating the development and evaluation of improved methods for handling potentially idiomatic MWEs in natural language. While there exist datasets for evaluating models' ability to identify idiomaticity~\cite{haagsma2020magpie,korkontzelos2013semeval,cook2008vnc,cordeiro2019unsupervised,garcia2021probing,shwartz2019still}, these are often not particularly suited to investigating a) the transfer learning capabilities across different data set-ups b) the performance of pre-trained contextualised models. 

The task consists of two subtasks: Subtask A, which is focused on the detection of idiomaticity, and Subtask B, which is focused on the representation of idiomaticity. In this work we are interested in the task of idiomaticity classification, since we wish to investigate how our models can identify idiomaticity in text without having to generate semantic similarity scores. As such, we restrict our attention to Subtask A. We also want to see how our models perform when MWEs in the test data are disjoint from those in the training data, as we argue this means the models cannot so easily leverage statistical information garnered from the training data, but must instead have some `knowledge' of idiomaticity in general. As such, we also restrict our attention to the zero-shot setting of the SemEval task. The dataset consists of three languages: English, Portuguese and Galician. In the training data there are 3,327 entries in English, and 1,164 entries in Portuguese. There is no Galician training (or development) data in the zero-shot setting, to test the ability of models at cross-lingual transfer. In the test set, there are 916 English, 713 Portuguese, and 713 Galician examples, and macro F1 score is used as an evaluation metric.

It should be noted that the dataset provided by \newcite{tayyarmadabushi-etal-2022-semeval} consists of four data splits: The training set, two development sets and the test set. Of the two development sets, the first - called the `dev' split - includes gold labels and the second - called the `eval' split - does not include gold labels but requires submission to the competition website. We report our results on the `eval' set to maintain consistency with the SemEval task.

\section{Methods} % Dylan
\label{section:methods}
In this section we detail our use of PET, iPET and BERTRAM for the task of idiomaticity detection.

\begin{table*}[ht!]
\begin{center}
\begin{tabular}{|c|c|c|c|}
      \hline
      Model & EN & PT & Overall\\
      \hline
      mBERT \small \cite{tayyar-madabushi-etal-2021-astitchinlanguagemodels-dataset} & 0.7420 &  0.5519 & 0.6871 \\
      \hline
      PET-all (10 labelled) & 0.4365 & 0.2901	& 0.4267 \\
      \hline
      PET-all (100 labelled) & 0.5908 & \textbf{0.5718} & 0.5888 \\
      \hline
      PET-all (1000 labelled) & \textbf{0.7820} & 0.5619	& \textbf{0.7164} \\
      \hline
      PET-P1 (1000 labelled) & 0.6386 & 0.5507 & 0.6278 \\
      \hline
      PET-P2 (1000 labelled) & 0.6905 & 0.5495 & 0.6607 \\
      \hline
      PET-P3 (1000 labelled) & 0.7493 & 0.5474 & 0.6981 \\
      \hline
      PET-P4 (1000 labelled) & 0.7441 & 0.5315 & 0.6860 \\
      \hline
      PET-P5 (1000 labelled) & 0.7551 & 0.5680 & 0.7032 \\
      \hline
      iPET (1000 labelled) [P1 \& P2] & 0.6701 & 0.5648 & 0.6522 \\
      \hline
\end{tabular}
\caption{\label{table:eval-results}The F1 Score (Macro) on the \textit{eval} set, broken down into each language, for each of the models. Highest score for each language (or overall) shown in bold.}
\end{center}
\end{table*}

\subsection{PET and iPET}
During our experiments with PET and it's variants, we define and test 5 Pattern Verbaliser Pairs, shown in table \ref{table:PVPs}. P1 and P2 are generic prompts which do not give the model much more information about the example, whereas P3, P4, and P5 include the whole idiom within the prompt. We hypothesise that this will allow the model to understand which part of the example it should be focusing on. Each of the patterns we define is in English, even when the example sentence and idiom are in Portuguese or Galician --- we will investigate the effect that this has on the final performance across the languages, as we hypothesise this may not have an impact given our use of a multilingual model.

For each PVP, we train a classification model using mBERT as the MLM. Furthermore, we train a standard PET model using all of the patterns. An iPET model is also trained, however to evaluate how using only generic prompts affects the results, we only train our iPET model using PVPs P1 and P2, for 2 iterations. Each of the model setups is trained 3 times using different random seeds, and the final distilled model is then used to produce the presented results. 

% With ADAPET, we, again, train a separate model for each of the PVPs and one model using all PVPs, with mBERT as the MLM.

Additionally, we investigate how the number of labelled examples affects the achieved performance for each of the model setups discussed. We train the models using 10, 100, and 1000 labelled examples separately, with the examples chosen randomly across English and Portuguese, but with the split of idiomatic and literal uses being kept at 50/50. The PET and iPET models then have access to 3,000 unlabelled examples to use within their training tasks.

We evaluate each model setup and labelled example set size combination on the \textit{eval} set, before choosing the best-performing combination for each PET variant to evaluate on the test set. The results from the \textit{eval} set can be seen in Table \ref{table:eval-results}. Here we see that PET-all trained on 1000 labelled examples performs best overall, beating the individual pattern models, a result also seen in the original paper \cite{schick-schutze-2021-exploiting}. The lack of example specific prompts causes iPET to perform poorly when compared to the individual task specific patterns, and when compared to the best PET-all model. The highest scoring PET model (PET-all) and our iPET model are evaluated on the test dataset in Section \ref{section:results}.

\begin{table*}[ht!]
\begin{center}
\begin{tabular}{|c|c|c|c|c|}
      \hline
      Model & EN & PT  & GL  & Overall\\
      \hline
      mBERT \small \cite{tayyarmadabushi-etal-2022-semeval} & 0.7070 & \textbf{0.6803} & 0.5065 & \textbf{0.6540} \\
      \hline
      BERTRAM & \textbf{0.7769} & 0.5017 & 0.4994 & 0.6455 \\
      \hline
      PET-all (10 labelled) & 0.5197 & 0.2634 & 0.2090 & 0.4128 \\
      \hline
      PET-all (100 labelled) & 0.6777 & 0.5014 & 0.4902 & 0.5694 \\
      \hline
      PET-all (1000 labelled) & 0.7281 & 0.6253 & \textbf{0.5110} & 0.6446 \\
      \hline
      iPET (1000 labelled) [P1 \& P2] & 0.6604 & 0.5676 & 0.4735 & 0.5879 \\
      \hline
\end{tabular}
\caption{\label{table:results}The F1 Score (Macro) on the \emph{test} set, broken down into each language, for each of the models. Highest score for each language (or overall) shown in bold.}
\end{center}
\end{table*}

\subsection{BERTRAM}

To evaluate the effect that improved idiom representations have on this idiom detection task, we use the same BERTRAM setup as presented in \newcite{phelps-semeval}, that was shown to give greatly improved performance over the baseline system for Subtask B, the task of representing idiomaticity. We use the same BERTRAM models: the English model presented in the original BERTRAM paper \cite{schick-schutze-2020-bertram}, and the Portuguese and Galician models that were trained for Subtask B from data in the CC100 corpus. Unlike the English BERTRAM model, \newcite{phelps-semeval} does not use one token approximation when training the Portuguese and Galician models. Embeddings for each of the idioms in the task datasets were generated with the appropriate BERTRAM model using 150 examples scraped from the CC100 dataset. 150 examples were chosen as this was shown to have the highest performance on Subtask B. It should be noted that the BERTRAM models were used to create representations of MWEs in the test set. While this does not require labelled data associated with MWEs (thus remaining a zero-shot task), it does require knowledge of which phrases need to have explicit representations created. 

As we have separate BERTRAM models for each language that are trained to mimic embeddings from single language BERT models, we split the system and data into English, Portuguese and Galician. The English model uses BERT base \cite{devlin-etal-2019-bert}, and is trained on the 3,327 English training examples found in the training set. The Portuguese model uses BERTimbau \cite{portugueseBERT}, and Galician uses BERTinho \cite{galicianBERT}, and as there is no Galician training data available, both are trained on the 1,164 Portuguese examples. Each model has the MWEs from the relevant language added to its embedding matrix. 

\section{Results and Discussion} % Dylan

\begin{table*}[ht]
\begin{center}
% \label{table:translations}
\begin{tabular}{|c|c|c|c|}
      \hline
      Language & Pattern & Literal Token & Idiom Token \\
      \hline
      EN & X. \_\_\_\_, [IDIOM] is literal. & yes & no \\
      \hline
      PT & X. \_\_\_\_, [IDIOM] é literal. & sim & não \\
      \hline
      GL & X. \_\_\_\_, [IDIOM] é literal. & si & non \\
      \hline
\end{tabular}
\caption{\label{table:translations} The translations of P4 into Portuguese and Galician}
 \end{center}
\end{table*}

\begin{table*}[ht]
\begin{center}
\begin{tabular}{|c|c|c|c|c|c|}
      \hline
      Model & Prompt Language & EN & PT  & GL  & Overall\\
      \hline
      mBERT \small \cite{tayyarmadabushi-etal-2022-semeval} & N/A & 0.7070 & 0.6803 & 0.5065 & 0.6540 \\
      \hline
      PET-P4 (1000 labelled) & EN & 0.7161 & 0.6373 & 0.5365 & 0.6581 \\
      \hline
      PET-P4 (1000 labelled) & PT & 0.6994 & 0.6260 & 0.4964 & 0.6283 \\
      \hline
      PET-P4 (1000 labelled) & GL & 0.7040 & 0.5997 & 0.5154 & 0.6279 \\
      \hline
\end{tabular}
\caption{\label{table:translation-results}The F1 Score (Macro) on the \emph{test} set, broken down into each language, for PET using prompts in each of the task languages.}
 \end{center}
\end{table*}

\label{section:results}

Table \ref{table:results} presents the results of our best PET-based models alongside our BERTRAM-based model on the test set, as well as the mBERT system presented in \cite{tayyarmadabushi-etal-2022-semeval}, for comparison. For each model we present the F1 macro score on the test set for each language, as well as the overall F1 macro score.

An increase in performance over mBERT by our BERTRAM model is seen for the English split, with the score on the Galician split not seeing a significant change. The overall score for BERTRAM is brought down by a much lower score on the Portuguese data, however, meaning no overall increase in performance is seen. A similar picture is seen for the PET-all (1000 examples) model, with a higher F1 score in both English and Galician, and a lower score in Portuguese, leading to an overall lower F1 score across the entire test dataset. As found on the example data, the iPET model which was only trained on the non-example specific prompts (P1 and P2) performs very poorly.

The significant boost from using BERTRAM on English seems to indicate that the improved representations also lead to better classification, despite the lacklustre performance on Galician and Portuguese. We believe that this drop in performance is either because one-token approximation was not used in creating the non-English BERTRAM models, or because mBERT, trained on all three languages simultaneously, is trained on more data than each of our monolingual models. This lack of training data does not affect our English model as there is a more training data in English than in Portuguese and none at all in Galician. We perform a language specific error analysis to explore the causes of this drop in performance (Section \ref{sec:error-analysis}).

 It is interesting to note that pre-trained language models can identify idiomaticity in a zero-shot and sample efficient context \emph{even when prior work has shown that they do not encode idiomaticity very well} \cite{garcia-etal-2021-probing}. We believe that this implies that, while these models do not encode idiomaticity, they encode enough related information to be able to \emph{infer} idiomaticity from relatively little data. 

Unsurprisingly, `highlighting' the phrase that is potentially idiomatic by adding the phrase to the pattern, as in patterns P3, P4 and P5 (see Table \ref{table:PVPs}), significantly improves a model's ability to identify idiomaticity, which is consistent with results presented by~\newcite{tayyar-madabushi-etal-2021-astitchinlanguagemodels-dataset}.

\paragraph{Research Questions} The results presented herein suggest that few-shot learning methods are indeed effective on the task of idiomaticity detection despite the lower accuracy on Portuguese and Galician. Similarly, our results support the conclusion that improved MWE representations does have an impact on improved detection. 

\subsection{Error Analysis} \label{sec:error-analysis}

The effectiveness of PET on the English split of the task suggests that pre-trained language models can effectively identify idiomatic MWEs in a sample efficient manner. However, the overall drop in performance on the task can be attributed to lower performance on non-English languages when compared to the results achieved by ~\newcite{tayyar-madabushi-etal-2021-astitchinlanguagemodels-dataset}.

One possibility for the decrease in performance is the use of English prompts across all the languages. This leads to the inputs for English examples being monolingual and the inputs for non-English examples to be multilingual, which may cause confusion in the output logits for the verbalizer tokens from which PET draws it's predictions.

To investigate this further we translate one of our patterns, P4, into both Portuguese and Galician and evaluate the performance on the entire \textit{test} split. P4 was chosen as it was one of the better performing patterns for English in our initial experiments (Table \ref{table:eval-results}), and was easily translated into the two languages. The translations can be seen in table \ref{table:translations}.

As shown in table \ref{table:translation-results}, the use of Portuguese and Galician prompts does not increase the performance in the respective language. For Portuguese the model with Portuguese prompts achieves 0.6260 F1 score compare to 0.6373 for that with English prompts. Galician shows similar results, with 0.5154 F1 score for the model with prompts in Galician and 0.5365 for that in English. 

Additionally, we use multilingual BERT which was trained on a lot more English training data than Portuguese or Galician language. To investigate the impact of this on our results, we extract only the Portuguese section of the training and test data and compare the performance of multilingual BERT with Portuguese BERT \cite{portugueseBERT}. Surprisingly, we find that the there isn't a significant difference between the performance of multilingual BERT and Portuguese BERT, with overall F1 (macro) scores of 0.4541 and 0.4621, respectively.

\section{Conclusions and Future work}
\label{section:conclusion}

This work presented our exploration of \emph{sample efficient} methods for idiomaticity detection, crucial given the infrequent occurrence of specific MWEs in natural language text. Our experiments show that these methods are extremely promising and have great potential. 

In future work, we intend to raucously evaluate and find solutions to the problem of lower performance on non-English test splits. We also intend to explore other variations of BERTRAM (e.g. one-token approximation) in bridging the performance gap between English and the other languages. 

As noted earlier, we show that pre-trained language models can identify idiomaticity in a zero-shot and sample efficient context \emph{even when prior work has shown that they do not encode idiomaticity very well}. As such, an important avenue of future exploration is the generalisation of these methods to develop models capable of identifying \emph{the notion of idiomaticity}, much like humans are able to grasp that certain phrases are clearly non-compositional. 

\section*{Acknowledgements}
This work is partially supported the Healthy Lifespan Institute (HELSI) at The University of Sheffield and is funded by the Engineering and Physical Sciences Research Council (EPSRC) [grant number EP/T517835/1]. This work was also partially supported by the UK EPSRC grant EP/T02450X/1 and the CDT in Speech and Language Technologies and their Applications funded by UKRI [grant number EP/S023062/1].

% \nocite{*}
\section{Bibliographical References}\label{reference}
%\label{main:ref}

\bibliographystyle{lrec2022-bib}
\bibliography{lrec2022-example}

%\section{Language Resource References}
%\label{lr:ref}
% \bibliographystylelanguageresource{lrec2022-bib}
% \bibliographylanguageresource{languageresource}

\end{document}